\documentclass[11pt,a4paper]{article}
\usepackage[hyperref]{acl2019}
\usepackage{times}
\usepackage{latexsym}

\usepackage{url}

\usepackage{amssymb}
\usepackage{amsfonts}
\usepackage{graphicx}
\usepackage{framed}
\usepackage{colortbl}
\usepackage{color}
\usepackage{multicol}
\usepackage{multirow}
\usepackage{makecell}
\usepackage{amsmath}
\usepackage{booktabs}
\usepackage{array}
\usepackage{arydshln}
\usepackage{paralist}

\allowdisplaybreaks[1] 

\definecolor{gray_low}{gray}{.9}
\definecolor{gray_mid}{gray}{.8}
\definecolor{gray_high}{gray}{.7}
\aclfinalcopy 
\def\aclpaperid{1933} 

\newcommand\lock{\includegraphics[width = 0.18cm]{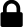}}
\newcommand\graylow{ \multicolumn{1}{>{\columncolor{gray_low}}c}}
\newcommand\graymid{ \multicolumn{1}{>{\columncolor{gray_mid}}c}}
\newcommand\grayhigh{ \multicolumn{1}{>{\columncolor{gray_high}}c}}

\title{Jointly Learning Semantic Parser and Natural Language Generator \\via Dual Information Maximization}

\author{Hai Ye$^\flat$ \ \ \ \ \ Wenjie Li$^\natural$ \ \ \ \ \ Lu Wang$^\sharp$ \\
$^\flat$ Information Sciences Institute, University of Southern California\\
$^\natural$ Department of Computing, The Hong Kong Polytechnic University\\
$^\sharp$ College of Computer and Information Science, Northeastern University\\
 \texttt{hye@usc.edu}, \texttt{cswjli@comp.polyu.edu.hk}, \texttt{luwang@ccs.neu.edu}}

\date{}

\begin{document}
\maketitle
\begin{abstract}
Semantic parsing aims to transform natural language~(NL) utterances into formal meaning representations~(MRs), whereas an NL generator achieves the reverse: producing a NL description for some given MRs. Despite this intrinsic connection, the two tasks are often studied separately in prior work. 
In this paper, we model the \textit{duality} of these two tasks via a joint learning framework, and demonstrate its effectiveness of boosting the performance on both tasks. 
%
Concretely, we propose the method of \emph{dual information maximization}~(\textsc{DIM}) to regularize the learning process, where \textsc{DIM} empirically maximizes the \emph{variational lower bounds} of expected joint distributions of NL and MRs. 
We further extend \textsc{DIM} to a semi-supervision setup~(\textsc{SemiDIM}), which leverages unlabeled data of both tasks. 
Experiments on three datasets of dialogue management and code generation (and summarization) show that performance on both semantic parsing and NL generation can be consistently improved by \textsc{DIM}, in both supervised and semi-supervised setups\footnote{Code for this paper is available at: \url{https://github.com/oceanypt/DIM}}. 
\end{abstract}

\section{Introduction}
Semantic parsing studies the task of translating natural language~(NL) utterances into formal meaning representations~(MRs)~\cite{DBLP:conf/aaai/ZelleM96,DBLP:conf/emnlp/TangM00}. NL generation models can be designed to learn the reverse: mapping MRs to their NL descriptions~\cite{DBLP:conf/naacl/WongM07}. 
Generally speaking, MR often takes a logical form that captures the semantic meaning, including $\lambda$-calculus~\cite{DBLP:conf/uai/ZettlemoyerC05,DBLP:conf/emnlp/ZettlemoyerC07}, Abstract Meaning Representation (AMR)~\cite{DBLP:conf/acllaw/BanarescuBCGGHK13,DBLP:conf/emnlp/MisraA16}, and general-purpose computer programs, such as Python~\cite{DBLP:conf/acl/YinN17} or SQL~\cite{DBLP:journals/corr/abs-1709-00103}. 
%
Recently, NL generation models have been proposed to automatically construct human-readable descriptions from MRs, for code summarization~\cite{DBLP:conf/iwpc/HuLXLJ18,allamanis2016convolutional,DBLP:conf/acl/IyerKCZ16} that predicts the function of code snippets, and for AMR-to-text generation~\cite{DBLP:conf/acl/GildeaWZS18,DBLP:conf/acl/KonstasIYCZ17,DBLP:conf/naacl/FlaniganDSC16}.

\begin{figure}[t]
\setlength{\abovecaptionskip}{0.2cm}
\begin{center}
\includegraphics[width=7cm]{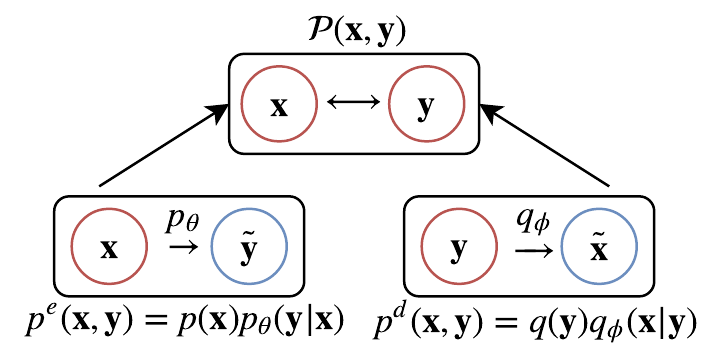}
\end{center}
   \caption{
   Illustration of our joint learning model. $\mathbf{x}$: NL; $\mathbf{y}$: MRs. $p_\theta (\mathbf{y}\vert \mathbf{x})$: semantic parser; $q_\phi (\mathbf{x}\vert \mathbf{y})$: NL generator.
   We model the duality of the two tasks by matching the joint distributions of $p^e(\mathbf{x}, \mathbf{y})$ (learned from semantic parser) and $p^d(\mathbf{x}, \mathbf{y})$ (learned from NL generator) to an underlying unknown distribution $\mathcal{P}(\mathbf{x}, \mathbf{y})$. 
   }
\label{fig:DIM}
\end{figure}

\begin{figure*}[t]
\setlength{\abovecaptionskip}{0.2cm}
\begin{center}
\includegraphics[width=\textwidth]{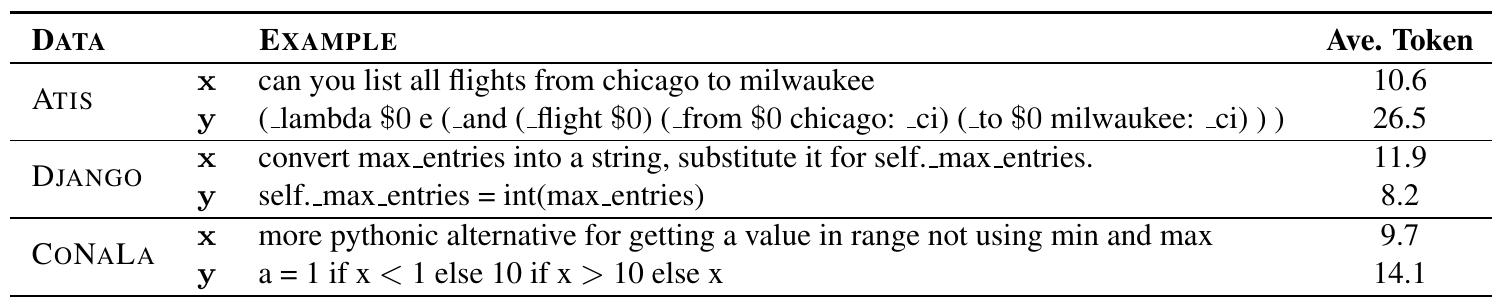}
\end{center}
   \caption{Sample natural language utterances and meaning representations from datasets used in this work: \textsc{Atis} for dialogue management; \textsc{Django}~\cite{DBLP:conf/kbse/OdaFNHSTN15} and \textsc{CoNaLa}~\cite{DBLP:conf/msr/YinDCVN08} for code generation and summarization.
   }
\label{tab:dataset}
\end{figure*}

%
%
Specifically, a common objective that semantic parsers aim to estimate is $p_\theta(\mathbf{y} | \mathbf{x})$, the conditional distribution  between NL input $\mathbf{x}$ and the corresponding MR output $\mathbf{y}$, as demonstrated in Fig.~\ref{fig:DIM}. 
Similarly, for NL generation from MRs, the goal is to learn a generator of $q_\phi(\mathbf{x} | \mathbf{y})$. %
As demonstrated in Fig.~\ref{tab:dataset}, there is a clear \emph{duality} between the two tasks, given that one task's input is the other task's output, and vice versa.
%
However, such duality remains largely unstudied, even though joint modeling has been demonstrated effective in various NLP problems, e.g. question answering and generation~\cite{DBLP:journals/corr/TangDQZ17}, machine translation between paired languages~\cite{DBLP:conf/nips/HeXQWYLM16}, as well as sentiment prediction and subjective text generation~\cite{DBLP:conf/icml/XiaQCBYL17}.



In this paper, we propose to {\it jointly model semantic parsing and NL generation by exploiting the interaction between the two tasks}. 
Following previous work on dual learning~\cite{DBLP:conf/icml/XiaQCBYL17}, we leverage the joint distribution $\mathcal{P}(\mathbf{x}, \mathbf{y})$ of NL and MR to represent the duality.  
Intuitively, as shown in Fig.~\ref{fig:DIM}, the joint distributions of $p^e(\mathbf{x}, \mathbf{y})=p(\mathbf{x}) p_\theta(\mathbf{y} | \mathbf{x})$, which is estimated from semantic parser, and $p^d(\mathbf{x}, \mathbf{y})= q(\mathbf{y}) q_\phi(\mathbf{x} | \mathbf{y})$, which is modeled by NL generator, are both expected to approximate $\mathcal{P}(\mathbf{x}, \mathbf{y})$, the unknown joint distribution of NL and MR. 

To achieve this goal, we propose \emph{dual information maximization}~(\textsc{DIM})~($\S$\ref{DIM}) to empirically optimize the \emph{variational lower bounds} of the expected joint distributions of $p^e(\mathbf{x}, \mathbf{y})$ and $p^d(\mathbf{x}, \mathbf{y})$. 
Concretely, the coupling of the two expected distributions is designed to capture the dual information, with both optimized via {variational approximation}~\cite{DBLP:conf/nips/BarberA03} inspired by \citet{DBLP:conf/nips/ZhangGGGLBD18}. Furthermore, combined with the supervised learning objectives of semantic parsing and NL generation, \textsc{DIM} bridges the two tasks within one joint learning framework by serving as a regularization term~($\S$\ref{sec:learning object}). 
Finally, we extend supervised \textsc{DIM} to semi-supervision setup~(\textsc{SemiDIM}), where unsupervised learning objectives based on unlabeled data are also optimized~($\S$\ref{SemiDIM}).

We experiment with three datasets from two different domains: \textsc{Atis} for dialogue management; \textsc{Django} and \textsc{CoNaLa} for code generation and summarization. 
Experimental results show that 
both the semantic parser and generator can be consistently improved with joint learning using \textsc{DIM} and \textsc{SemiDIM}, compared to competitive comparison models trained for each task separately. 

Overall, we have the following contributions in this work:
 \begin{compactitem}
     \item[$\bullet$]We are the first to jointly study semantic parsing and natural language generation by exploiting the \emph{duality} between the two tasks;
     \item[$\bullet$]We propose \textsc{DIM} to capture the duality and adopt variational approximation to maximize the dual information;
    \item[$\bullet$]We further extend supervised \textsc{DIM} to semi-supervised setup~(\textsc{SemiDIM}). 
 \end{compactitem}

\section{Problem Formulation}
\subsection{Semantic Parsing and NL Generation}\label{sec:parser and generator}
Formally, the task of semantic parsing is to map the input of NL utterances $\mathbf{x}$ to the output of structured MRs $\mathbf{y}$, and NL generation learns to generate NL from MRs. 

\noindent{\textbf{Learning Objective.}} \ \ Given a labeled dataset $\mathbb{L} = \{\langle \mathbf{x}_i, \mathbf{y}_i \rangle \}$, we aim to learn a semantic parser~($\mathbf{x} \to \mathbf{y}$) by estimating the conditional distribution $p_\theta(\mathbf{y}| \mathbf{x})$, parameterized by $\theta$, and an NL generator~($\mathbf{y} \to \mathbf{x}$) by modeling $q_\phi(\mathbf{x}|\mathbf{y})$, parameterized by $\phi$. The learning objective for each task is shown below: 
\begingroup
\setlength\abovedisplayskip{4pt}\setlength\belowdisplayskip{4pt}
\begin{align}
    \mathcal{L}_{\text{parser}} = \mathbb{E}_{\langle \mathbf{x}, \mathbf{y} \rangle} \lbrack \log p_\theta(\mathbf{y} | \mathbf{x}) \rbrack \label{obj:parser} \\
    \mathcal{L}_{\text{gen.}} = \mathbb{E}_{\langle \mathbf{x}, \mathbf{y} \rangle } \lbrack \log q_\phi(\mathbf{x} | \mathbf{y}) \rbrack \label{obj:gen}
\end{align}
\endgroup

\noindent{\textbf{Frameworks.}} \ \ Sequence-to-sequence (seq2seq) models have achieved competitive results on both semantic parsing and generation~\cite{dong-lapata:2016:P16-1,DBLP:conf/iwpc/HuLXLJ18}, and without loss of generality, we adopt it as the basic framework for both tasks in this work. 
Specifically, for both $p_\theta(\mathbf{y} | \mathbf{x})$ and $q_\phi(\mathbf{x} | \mathbf{y})$, we use a two-layer bi-directional LSTM (bi-LSTM) as the encoder and another one-layer LSTM as the decoder with attention mechanism~\cite{DBLP:conf/emnlp/LuongPM15}. Furthermore, we leverage pointer network~\cite{DBLP:conf/nips/VinyalsFJ15} to copy tokens from the input to handle out-of-vocabulary (OOV) words. The structured MRs are linearized for the sequential encoder and decoder. 
More details of the parser and the generator can be found in Appendix~\ref{app:framework}. Briefly speaking, our models differ from existing work as follows:
\begin{compactitem}
\item[\textbf{\textsc{Parser}:}] Our architecture is similar to the one proposed in~\newcite{DBLP:conf/acl/JiaL16} for semantic parsing; 
\item[\textbf{\textsc{Generator}:}] Our model improves upon the \textsc{DeepCom} coder summarization system~\cite{DBLP:conf/iwpc/HuLXLJ18} by: 1) replacing LSTM with bi-LSTM for the encoder to better model context, and 2) adding copying mechanism. 
\end{compactitem}


\subsection{Jointly Learning Parser and Generator}\label{sec:learning object}
Our joint learning framework is designed to model the duality between a parser and a generator. 
To incorporate the duality into our learning process, we design the framework to encourage the expected joint distributions $p^e(\mathbf{x}, \mathbf{y})$ and $p^d(\mathbf{x}, \mathbf{y})$ to both approximate the unknown joint distribution of $\mathbf{x}$ and $\mathbf{y}$~(shown in Fig.~\ref{fig:DIM}). 
To achieve this, we introduce \emph{dual information maximization}~(\textsc{DIM}) to empirically optimize the \emph{variational lower bounds} of both $p^e(\mathbf{x}, \mathbf{y})$ and $p^d(\mathbf{x}, \mathbf{y})$, in which the coupling of expected distributions is captured as dual information~(detailed in $\S$\ref{dual inf}) and will be maximized during learning.

Our joint learning objective takes the form of:
\begingroup
\setlength\abovedisplayskip{4pt}\setlength\belowdisplayskip{4pt}
\begin{equation}
     \max\limits_{\theta, \phi} \mathcal{L}(\theta, \phi) = \mathcal{L}_{\text{parser}} + \mathcal{L}_{\text{gen.}} + \lambda \cdot \mathcal{L}_{\text{DIM}}(\theta, \phi)
     \label{func:joint}
\end{equation}
\endgroup
$\mathcal{L}_{\text{DIM}}$ is the variational lower bound of the two expected joint distributions, specifically, 
\begingroup
\setlength\abovedisplayskip{4pt}\setlength\belowdisplayskip{4pt}
\begin{equation}
    \mathcal{L}_{\text{DIM}} = \mathcal{L}^e_{\text{DIM}}~\text{\small{(Eq.~\ref{dim:lower bound})}} + \mathcal{L}^d_{\text{DIM}}~\text{\small{(Eq.~\ref{eq:lower_bound_bw})}}
\end{equation}
\endgroup
where $\mathcal{L}^e_{\text{DIM}}$ and $\mathcal{L}^d_{\text{DIM}}$ are the lower bounds over $p^e(\mathbf{x}, \mathbf{y})$ and $p^d(\mathbf{x}, \mathbf{y})$ respectively. 
The hyper-parameter $\lambda$ trades off between supervised objectives and dual information learning. 
With the objective of Eq.~\ref{func:joint}, we jointly learn a parser and a generator, as well as maximize the dual information between the two. $\mathcal{L}_{\text{DIM}}$ serves as a regularization term to influence the learning process, whose detailed algorithm is described in $\S$\ref{DIM}. 

Our method of DIM is model-independent. If the learning objectives for semantic parser and NL generator are subject to Eq.~\ref{obj:parser} and Eq.~\ref{obj:gen}, we can always adopt DIM to conduct joint learning. Out of most commonly used seq2seq models for the parser and generator, more complex tree and graph structures have been adopted to model MRs~\cite{dong-lapata:2016:P16-1,DBLP:conf/acl/GildeaWZS18}. In this paper, without loss of generality, we study our joint-learning method on the widely-used seq2seq frameworks mentioned above~($\S$\ref{sec:parser and generator}). 





\section{Dual Information Maximization}\label{DIM}
In this section, we first introduce dual information in $\S$\ref{dual inf}, followed by its maximization ($\S$\ref{max dual inf}). $\S$\ref{SemiDIM} discusses its extension with semi-supervision.

\subsection{Dual Information}\label{dual inf}
As discussed above, we treat semantic parsing and NL generation as the dual tasks and exploit the duality between the two tasks for our joint learning. 
With conditional distributions $p_\theta(\mathbf{y} | \mathbf{x})$ for the parser and $q_\phi(\mathbf{x} | \mathbf{y})$ for the generator, the joint distributions of $p^e(\mathbf{x}, \mathbf{y})$ and $p^d(\mathbf{x}, \mathbf{y})$ can be estimated as $p^e(\mathbf{x}, \mathbf{y})=p(\mathbf{x}) p_\theta(\mathbf{y} | \mathbf{x})$ and $p^d(\mathbf{x}, \mathbf{y})=q(\mathbf{y}) q_\phi(\mathbf{x} | \mathbf{y})$, where $p(\mathbf{x})$ and $q(\mathbf{y})$ are marginals. 
The dual information  $I_{p^{e,d}(\mathbf{x}, \mathbf{y})}$ between the two distributions is defined as follows:
\begingroup
\begin{equation}
\setlength\abovedisplayskip{4pt}\setlength\belowdisplayskip{4pt}
\begin{split}
    I_{p^{e,d}(\mathbf{x}, \mathbf{y})} = & \ I_{p^e(\mathbf{x}, \mathbf{y})} + I_{p^d(\mathbf{x}, \mathbf{y})} \triangleq \\  \mathbb{E}_{p^e(\mathbf{x}, \mathbf{y})}   \log  p^e(\mathbf{x}, & \mathbf{y})  + \mathbb{E}_{p^d(\mathbf{x}, \mathbf{y})} \log p^d(\mathbf{x}, \mathbf{y})
   \label{func:dual-info}
\end{split}
\end{equation}
\endgroup
which is the combination of the two joint distribution expectations. 

To leverage the duality between the two tasks, we aim to drive the learning of the model parameters $\theta$ and $\phi$ via optimizing $I_{p^{e,d}(\mathbf{x}, \mathbf{y})}$, so that the expectations of joint distributions $p^e(\mathbf{x}, \mathbf{y})$ and $p^d(\mathbf{x}, \mathbf{y})$ will be both maximized and approximate the latent joint distribution $\mathcal{P}(\mathbf{x}, \mathbf{y})$, whose procedure is similar to the joint distribution matching~\cite{DBLP:conf/nips/GanCWPZLLC17}. 
By exploiting the inherent probabilistic connection between the two distributions, we hypothesize that it would enhance the learning of both tasks on parsing $p_\theta(\mathbf{y} | \mathbf{x})$ and generation $q_\phi(\mathbf{x} | \mathbf{y})$. Besides, to approach the same distribution $\mathcal{P}(\mathbf{x}, \mathbf{y})$, the expected joint distributions can learn to be close to each other, making the dual models coupled.

\begin{figure*}[t]
\setlength{\abovecaptionskip}{0.2cm}
\begin{center}
\includegraphics[width=\textwidth]{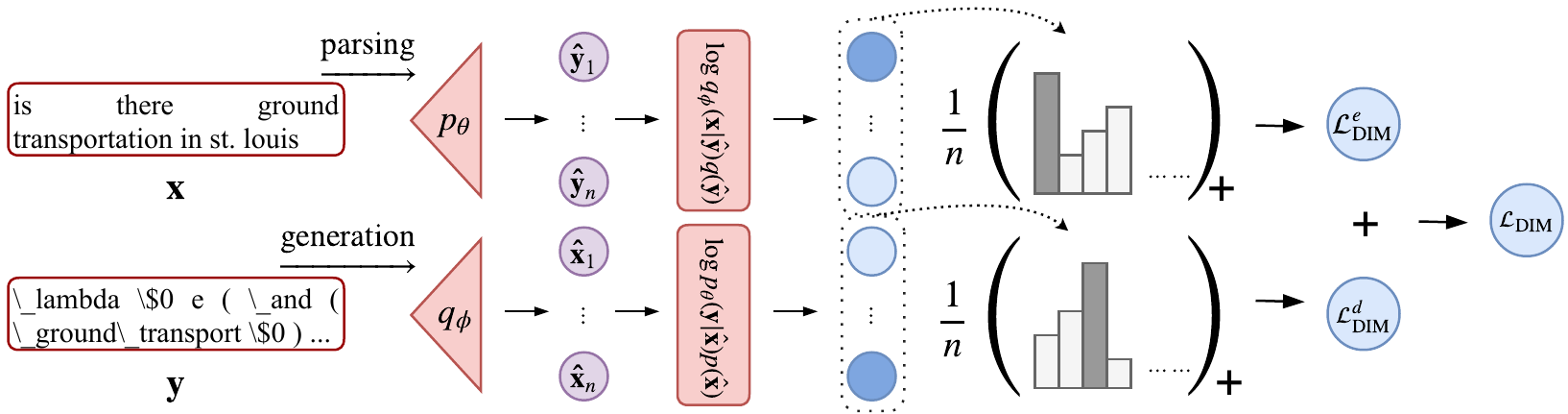}
\end{center}
   \caption{The pipeline of calculating lower bounds. We firstly use the parser or generator to sample MR or NL targets, then the sampled candidates go through the dual model and a language model to obtain the lower bounds.}
\label{fig:DIM-cal}
\end{figure*}

\subsection{Maximizing Dual Information}\label{max dual inf}
Here, we present the method for optimizing $I_{p^e(\mathbf{x}, \mathbf{y})}$, which can also be applied to $I_{p^d(\mathbf{x}, \mathbf{y})}$. 
In contrast to the parameter sharing techniques in most multi-task learning work~\cite{DBLP:journals/jmlr/CollobertWBKKK11,DBLP:journals/jmlr/AndoZ05}, parameter $\theta$ for the parser and parameter $\phi$ for generator are independent in our framework. In order to jointly train the two models and bridge the learning of $\theta$ and $\phi$, 
during the optimization of $I_{p^e(\mathbf{x}, \mathbf{y})}$, where the parser is the primal model, we utilize the distributions of the dual task (i.e. the generator) to estimate $I_{p^e(\mathbf{x}, \mathbf{y})}$. 
In this way, $\theta$ and $\phi$ can be both improved during the update of $I_{p^e(\mathbf{x}, \mathbf{y})}$. 
Specifically, we rewrite $\mathbb{E}_{p^e(\mathbf{x}, \mathbf{y})} \log p^e(\mathbf{x}, \mathbf{y})$ as $\mathbb{E}_{p^e(\mathbf{x}, \mathbf{y})} \log p^e(\mathbf{y})p^e( \mathbf{x} | \mathbf{y})$, where $p^e(\mathbf{y})$ and $p^e(\mathbf{x} | \mathbf{y})$ are referred as the dual task distributions. However, the direct optimization for this objective is impractical since both $p^e(\mathbf{y})$ and $p^e(\mathbf{x}|\mathbf{y})$ are unknown. Our solution is detailed below.

\noindent{\textbf{Lower Bounds of Dual Information.}}\ \ 
To provide a principled approach of optimizing $I_{p^e(\mathbf{x}, \mathbf{y})}$, inspired by \citet{DBLP:conf/nips/ZhangGGGLBD18}, we follow~\citet{DBLP:conf/nips/BarberA03} to adopt \emph{variational approximation} to deduce its lower bound and instead maximize the lower bound. The lower bound deduction process is as following:
\begingroup
\setlength\abovedisplayskip{4pt}\setlength\belowdisplayskip{4pt}
\begin{align}
\nonumber \mathbb{E}_{p^e(\mathbf{x}, \mathbf{y})} \log p^e(\mathbf{x}, \mathbf{y}) = \mathbb{E}_{p^e(\mathbf{x},\mathbf{y})} \log p^e(\mathbf{x} | \mathbf{y}) p^e(\mathbf{y}) \\
\nonumber =  \mathbb{E}_{p^e(\mathbf{x},\mathbf{y})} \log q_\phi(\mathbf{x}|\mathbf{y}) + \mathbb{E}_{p^e(\mathbf{x},\mathbf{y})} \log q(\mathbf{y}) \\
\nonumber + \mathbb{E}_{p^e(\mathbf{y})} \big [ \text{KL}(p^e(\mathbf{x}|\mathbf{y}) \|q_\phi(\mathbf{x}|\mathbf{y})) \big ] \\
\nonumber + \mathbb{E}_{p^e(\mathbf{x}|\mathbf{y})} \big [ \text{KL}(p^e(\mathbf{y}) \| q(\mathbf{y})) \big ] \\
\geqslant \mathbb{E}_{p^e(\mathbf{x},\mathbf{y})} \big[\log q_\phi(\mathbf{x}|\mathbf{y})  + \log q(\mathbf{y}) \big ]  = \mathcal{L}^e_{\text{DIM}}(\theta, \phi)
\label{dim:lower bound}
\end{align}
\endgroup
where $\text{KL}(\cdot \| \cdot)(\geqslant 0)$ is the Kullback-Leibler~(KL) divergence. 
Therefore, to maximize $I_{p^e(\mathbf{x}, \mathbf{y})}$, we can instead maximize its lower bound of $\mathcal{L}^e_{\text{DIM}}$. 
$\mathcal{L}^e_{\text{DIM}}$ is learned by using $q_\phi(\mathbf{x}|\mathbf{y})$ and $q(\mathbf{y})$ which approximate $p^e(\mathbf{x}|\mathbf{y})$ and $p^e(\mathbf{y})$. Besides, the lower bound of $\mathcal{L}^e_{\text{DIM}}$ is the function of $\theta$ and $\phi$, so in the process of learning $\mathcal{L}^e_{\text{DIM}}$, the parser and generator can be both optimized. 

As illustrated in Fig.~\ref{fig:DIM-cal}, in the training process, to calculate the lower bound of $\mathcal{L}^e_{\text{DIM}}$, we first use the being-trained parser to sample MR candidates for a given NL utterance. The sampled MRs then go through the generator and a marginal model~(i.e., a language model of MRs) to obtain the final lower bound.  

To learn the lower bound of $\mathcal{L}^e_{\text{DIM}}$, we provide the following method to calculate its gradients:

\noindent{\textbf{Gradient Estimation.}}\ \ We adopt Monte Carlo samples using  the REINFORCE policy~\cite{DBLP:journals/ml/Williams92} to approximate the gradient of $\mathcal{L}^e_{\text{DIM}}(\theta, \phi)$ with regard to $\theta$: 
\begingroup
\begin{equation}
\setlength\abovedisplayskip{4pt}\setlength\belowdisplayskip{4pt}
    \begin{split}
          \nabla_{\theta}& \mathcal{L}^e_{\text{DIM}}(\theta, \phi)  =   \mathbb{E}_{p_\theta(\mathbf{y}|\mathbf{x})} \nabla_{\theta} \log p_\theta(\mathbf{y}|\mathbf{x})  \\
 &\ \ \ \ \ \ \ \ \ \ \ \ \ \ \ \ \cdot [\log q_\phi(\mathbf{x}|\mathbf{y}) + \log q(\mathbf{y}) - \mathbf{b}] \\
  & = \mathbb{E}_{p_\theta(\mathbf{y}|\mathbf{x})} \nabla_{\theta} \log p_\theta(\mathbf{y}|\mathbf{x}) \cdot l(\mathbf{x}, \mathbf{y}; \phi) \\
     & \approx \frac{1}{|\mathcal{S}|} \sum \limits_{\hat{\mathbf{y}}_i \in \mathcal{S}} \nabla_{\theta} \log p_\theta(\hat{\mathbf{y}}_i|\mathbf{x}) \cdot l(\mathbf{x}, \hat{\mathbf{y}}_i; \phi)
    \label{grad:1}
    \end{split}
\end{equation}
\endgroup
$l(\mathbf{x}, \mathbf{y};\phi)$ can be seen as the learning signal from the dual model, which is similar to the \emph{reward} in reinforcement learning algorithms~\cite{DBLP:conf/acl/GuuPLL17,DBLP:journals/corr/PaulusXS17}. 
To handle the high-variance of learning signals, we adopt the baseline function $\mathbf{b}$ by empirically averaging the signals to stabilize the learning process~\cite{DBLP:journals/ml/Williams92}. 
With prior $p_\theta{(\cdot | \mathbf{x})}$, we use beam search to generate a pool of MR candidates ($\mathbf{y}$), denoted as $\mathcal{S}$, for the input of $\mathbf{x}$. 

The gradient with regard to $\phi$ is then calculated as:
\begingroup
\begin{equation}
\setlength\abovedisplayskip{4pt}\setlength\belowdisplayskip{4pt}
    \begin{split}
        \nabla_{\phi} \mathcal{L}^e_{\text{DIM}}(\theta, \phi)& = \mathbb{E}_{p_\theta(\mathbf{y}|\mathbf{x})} \nabla_{\phi} \log q_\phi(\mathbf{x}|\mathbf{y}) \\
  & \approx \frac{1}{|\mathcal{S}|} \sum \limits_{\hat{\mathbf{y}}_i \in \mathcal{S}} \nabla_{\phi} \log q_\phi(\mathbf{x}|\hat{\mathbf{y}}_i)
  \label{grad:2}
    \end{split}
\end{equation}
\endgroup

The above maximization procedure for $\mathcal{L}^e_{\text{DIM}}$ is analogous to the EM algorithm: 
\begin{compactitem}
\item[Step 1:] Freeze $\phi$ and find the optimal $\theta^*$ $=$ $\arg \max_{\theta} \mathcal{L}^e_{\text{DIM}}(\theta, \phi)$ with Eq.~\ref{grad:1}; 
\item[Step 2:] Based on Eq.~\ref{grad:2}, with freezing $\theta^*$, find the optimal $\phi^* = \arg \max_{\phi} \mathcal{L}^e_{\text{DIM}}(\theta, \phi)$. 
\end{compactitem}
The two steps are repeated until convergence.

According to the gradient estimation in Eq.~\ref{grad:1}, when updating $\theta$ for the parser, we receive the \emph{learning signal} $l(\mathbf{x}, \mathbf{y};\phi)$ from the generator, and this learning signal can be seen as a \emph{reward} from the generator: if parser $p_\theta(\mathbf{y} | \mathbf{x})$ predicts high-quality MRs, the reward will be high; otherwise, the reward is low. 
This implies that the generator guides the parser to generate high-quality MRs, through which the lower bound for the expected joint distribution gets optimized. This also applies to the situation when we treat the generator as the primal model and the parser as the dual model.

The lower bound of $I_{p^d(\mathbf{x},\mathbf{y})}$ can be calculated in a similar way: 
\begingroup
\setlength\abovedisplayskip{4pt}\setlength\belowdisplayskip{4pt}
\begin{multline}
 \mathbb{E}_{p^d(\mathbf{x}, \mathbf{y})} \log p^d(\mathbf{x}, \mathbf{y}) \\
\geqslant \mathbb{E}_{p^d(\mathbf{x},\mathbf{y})}\big [\log p_\theta(\mathbf{y}|\mathbf{x}) + \log p(\mathbf{x})\big ] = \mathcal{L}^d_{\text{DIM}}(\theta, \phi)
 \label{eq:lower_bound_bw}
\end{multline}
\endgroup
which can be optimized the same way as in Eqs.~\ref{grad:1} and~\ref{grad:2} for estimating the gradients for $\mathcal{L}^d_{\text{DIM}}$. 


\noindent{\textbf{Marginal Distributions.}}\ \ To obtain the marginal distributions $p(\mathbf{x})$ and $q(\mathbf{y})$, we separately train an LSTM-based language model~\cite{DBLP:conf/interspeech/MikolovKBCK10} for NL and MR respectively, on each training set. 
Structured MRs are linearized into sequences for the sequential encoder and decoder in seq2seq models. Details on learning marginal distributions can be found in Appendix~\ref{app:marginal}. 

\noindent{\textbf{Joint Learning Objective.}}\ \ 
Our final joint learning objective becomes:
\begingroup
\setlength\abovedisplayskip{4pt}\setlength\belowdisplayskip{4pt}
\begin{multline}
\max\limits_{\theta, \phi} \mathcal{J} = \sum\limits_{\langle \mathbf{x}, \mathbf{y} \rangle\in \mathbb{L}} \Big( \log p_\theta(\mathbf{y} | \mathbf{x}) + \log q_\phi(\mathbf{x} | \mathbf{y})\\
+ \lambda \sum\limits_{\hat{\mathbf{y}}_i \sim p_\theta(\cdot | \mathbf{x})}\log q_\phi(\mathbf{x} | \hat{\mathbf{y}}_i) + \log q(\hat{\mathbf{y}}_i) \\
+ \lambda \sum\limits_{\hat{\mathbf{x}}_i \sim q_\phi(\cdot | \mathbf{y})}\log p_\theta(\mathbf{y} | \hat{\mathbf{x}}_i) + \log p(\hat{\mathbf{x}}_i) \Big)
    \label{func:final:joint}
\end{multline}
\endgroup
According to this learning objective, after picking up a data pair $\langle \mathbf{x}, \mathbf{y} \rangle$, we will firstly calculate the supervised learning loss, then we sample MR candidates and NL samples using prior $p_\theta(\cdot | \mathbf{x})$ and $q_\phi(\cdot | \mathbf{y})$ respectively to obtain the corresponding lower bounds over $I_{p^e(\mathbf{x}, \mathbf{y})}$ and $I_{p^d(\mathbf{x}, \mathbf{y})}$. 

\subsection{Semi-supervised DIM (\textsc{SemiDIM})}\label{SemiDIM}
We further extend \textsc{DIM} with semi-supervised learning. We denote the unlabeled NL dataset as $\mathbb{U}_\mathbf{x} = \{\mathbf{x}_i\}$ and the unlabeled MR dataset as $\mathbb{U}_\mathbf{y} = \{\mathbf{y}_i\}$. To leverage $\mathbb{U}_{\mathbf{x}}$, we maximize the unlabeled objective $\mathbb{E}_{\mathbf{x} \sim \mathbb{U}_{\mathbf{x}}}\log p(\mathbf{x})$. 
Our goal is to involve model parameters in the optimization process of $\mathbb{E}_{\mathbf{x} \sim \mathbb{U}_{\mathbf{x}}}\log p(\mathbf{x})$, so that the unlabeled data can facilitate parameter leanring. 

\noindent{\textbf{Lower Bounds of Unsupervised Objective.}} \ \ 
The lower bound of $\mathbb{E}_{\mathbf{x} \sim \mathbb{U}_{\mathbf{x}}}\log p(\mathbf{x})$ is as follows, using the deduction in Ineq.~\ref{dim:lower bound}: 
\begingroup
\setlength\abovedisplayskip{4pt}\setlength\belowdisplayskip{4pt}
\begin{multline}
    \mathbb{E}_{\mathbf{x} \sim \mathbb{U}_{\mathbf{x}}}\log p(\mathbf{x}) \\ 
    \geq \mathbb{E}_{\mathbf{x} \sim \mathbb{U}_{\mathbf{x}}, \mathbf{y} \sim p_\theta(\cdot | \mathbf{x})}\log p(\mathbf{x})p_\theta(\mathbf{y} | \mathbf{x}) \\ 
    \geq \mathbb{E}_{\mathbf{x} \sim \mathbb{U}_{\mathbf{x}}, \mathbf{y} \sim p_\theta(\cdot | \mathbf{x})} \big[ \log q_\phi(\mathbf{x} | \mathbf{y}) + q(\mathbf{y}) \big]
    \label{semi:lower:bound}
\end{multline}
\endgroup
Comparing Ineq.~\ref{semi:lower:bound} to Ineq.~\ref{dim:lower bound}, we can see that the unsupervised objective $\mathbb{E}_{\mathbf{x} \sim \mathbb{U}_{\mathbf{x}}}\log p(\mathbf{x})$ and $I_{p^e(\mathbf{x},\mathbf{y})}$ share the same lower bound, so that the same optimization method from Eq.~\ref{grad:1} and Eq.~\ref{grad:2} can be utilized for learning the lower bound over $\mathbb{E}_{\mathbf{x} \sim \mathbb{U}_{\mathbf{x}}}\log p(\mathbf{x})$.

\noindent{\textbf{Analysis.}} \ \ The lower bound of the unsupervised objective $\mathbb{E}_{\mathbf{x} \sim \mathbb{U}_{\mathbf{x}}}\log p(\mathbf{x})$ is a function of $\theta$ and $\phi$. Therefore, updating this unsupervised objective will jointly optimize the parser and the generator. From the updating algorithm in Eq.~\ref{grad:1}, we can see that the parser $p_\theta(\mathbf{y} | \mathbf{x})$ is learned by using pseudo pair $(\mathbf{x}, \hat{\mathbf{y}})$ where $\hat{\mathbf{y}}$ is sampled from $p_\theta(\cdot | \mathbf{x})$. This updating process resembles the popular semi-supervised learning algorithm of \textbf{\emph{self-train}} that predicts pseudo labels for unlabeled data~\cite{lee2013pseudo} and then attaches the predicted labels to the unlabeled data as additional training data. 
In our algorithm, the pseudo sample $(\mathbf{x}, \hat{\mathbf{y}})$ will be weighted by the learning signal $l(\mathbf{x}, \hat{\mathbf{y}};\phi)$, which decreases the impact of low-quality pseudo samples. Furthermore, from Eq.~\ref{grad:2}, the generator $q_\phi(\mathbf{x} | \mathbf{y})$ is updated using the pseudo sample $(\mathbf{x}, \hat{\mathbf{y}})$, which is similar to the semi-supervised learning method of \textbf{\emph{back-boost}} that is widely used in Neural Machine Translation for low-resource language pairs~\cite{DBLP:conf/acl/SennrichHB16}. Given the target-side corpus, back-boost generates the pseudo sources to construct pseudo samples, which is added for model training. 

Similarly, to leverage the unlabeled data $\mathbb{U}_{\mathbf{y}}$ for semi-supervised learning, following Ineq.~\ref{semi:lower:bound}, we could also have the lower bound for $\mathbb{E}_{\mathbf{y} \sim \mathbb{U}_{\mathbf{y}}}\log p(\mathbf{y})$ as following, 
\begingroup
\setlength\abovedisplayskip{4pt}\setlength\belowdisplayskip{4pt}
\begin{multline}
    \mathbb{E}_{\mathbf{y} \sim \mathbb{U}_{\mathbf{y}}}\log p(\mathbf{y}) \\ 
    \geq \mathbb{E}_{\mathbf{y} \sim \mathbb{U}_{\mathbf{y}}, \mathbf{x} \sim q_\phi(\cdot | \mathbf{y})} \big[ \log p_\theta(\mathbf{y} | \mathbf{x}) + p(\mathbf{x}) \big]
    \label{semi:lower:bound:2}
\end{multline}
\endgroup
which is the same as the lower bound of $I_{p^d(\mathbf{x}, \mathbf{y})}$. 

\smallskip
\noindent{\textbf{Semi-supervised Joint Learning Objective.}} \ \ From the above discussions, we can deduce the lower bounds for the unsupervised objectives to be the same as the lower bounds of the dual information. We thus have the following semi-supervised joint-learning objective:
\begingroup
\setlength\abovedisplayskip{4pt}\setlength\belowdisplayskip{4pt}
\begin{align}
\nonumber \max\limits_{\theta, \phi} \mathcal{J} = \sum\limits_{\langle \mathbf{x}, \mathbf{y} \rangle\in \mathbb{L}} \big( \log p_\theta(\mathbf{y} | \mathbf{x}) + \log q_\phi(\mathbf{x} | \mathbf{y}) \big ) \\
\nonumber + \lambda \sum\limits_{\mathbf{x} \sim \mathcal{D}_\mathbf{x}, \hat{\mathbf{y}}_i \sim p_\theta(\cdot | \mathbf{x})}\big( \log q_\phi(\mathbf{x} | \hat{\mathbf{y}}_i) + \log q(\hat{\mathbf{y}}_i) \big ) \\
+ \lambda \sum\limits_{\mathbf{y} \sim \mathcal{D}_\mathbf{y}, \hat{\mathbf{x}}_i \sim q_\phi(\cdot | \mathbf{y})}\big ( \log p_\theta(\mathbf{y} | \hat{\mathbf{x}}_i) + \log p(\hat{\mathbf{x}}_i) \big )
    \label{func:final:semi-joint}
\end{align}
\endgroup
where $\mathcal{D}_\mathbf{x} = \mathbb{U}_\mathbf{x}\cup \mathbb{L}_\mathbf{x}$ and $\mathcal{D}_\mathbf{y} = \mathbb{U}_\mathbf{y}\cup \mathbb{L}_\mathbf{y}$. 
In this work, we weight the dual information and unsupervised objectives equally for simplicity, so the lower bounds over them are combined for joint optimization. We combine the labeled and unlabeled data to calculate the lower bounds to optimize the variational lower bounds of dual information and unsupervised objectives.


\begin{table}[]
\setlength{\abovecaptionskip}{0.2cm}
\centering
 \fontsize{9}{11.5}\selectfont
\begin{tabular}{p{1.3cm}p{.8cm}<{\centering} p{.8cm}<{\centering} p{.8cm}<{\centering} p{1cm}<{\centering}}\toprule[.5pt]
\scshape{Data}   & Train & Valid & Test  & All\\ \hline
\scshape{Atis}   & 4,480   & 480     & 450   & 5,410\\
\scshape{Django} & 16,000  & 1,000   & 1,805 & 18,805   \\
\scshape{CoNaLa} & 90,000  & 5,000   & 5,000 & 100,000 \\ \toprule[.5pt]
\end{tabular}
\caption{Statistics of datasets used for evaluation. Around 500K additional samples of low confidence from \textsc{CoNaLa} are retained for model pre-training.}
\label{tab:sta.dataset}
\end{table}

\section{Experiments}
\subsection{Datasets}
Experiments are conducted on three datasets with sample pairs shown in Fig.~\ref{tab:dataset}: one for dialogue management which studies semantic parsing and generation from $\lambda$-calculus~\cite{DBLP:conf/emnlp/ZettlemoyerC07}~(\textsc{Atis}) and two for code generation and summarization~(\textsc{Django}, \textsc{CoNaLa}). 

\noindent{$\text{\textbf{\scshape{Atis}.}}$}\ \ 
This dataset has 5,410 pairs of queries~(NL) from a flight booking system and corresponding $\lambda$-calculus representation~(MRs). 
The anonymized version from \citet{dong-lapata:2016:P16-1} is used. 

\noindent{$\text{\textbf{\scshape{Django}}.}$} \ \ It contains 18,805 lines of Python code snippets~\cite{DBLP:conf/kbse/OdaFNHSTN15}. Each snippet is annotated with a piece of human-written pseudo code. Similar to \citet{DBLP:conf/acl/YinN17}, we replace strings separated by quotation marks with indexed \emph{place\_holder} in NLs and MRs. 

\noindent{$\text{\textbf{\scshape{CoNaLa}}.}$} \ \ This is another Python-related corpus containing 598,237 intent/snippet pairs that are automatically mined from Stack Overflow~\cite{DBLP:conf/msr/YinDCVN08}. Different from \textsc{Django}, the intent in \textsc{CoNaLa} is mainly about the question on a specific topic instead of pseudo code. 
The full dataset contains noisy aligned pairs, and we keep the top 100,000 pairs of highest confidence scores for experiment and the rest for model pre-training.

For $\text{\scshape{Django}}$ and $\text{\scshape{CoNaLa}}$, the NL utterances are lowercased and tokenized and the tokens in code snippets are separated with space. Statistics of the datasets are summarized in Table~\ref{tab:sta.dataset}.

\subsection{Experimental Setups}
\noindent{\textbf{Joint-learning Setup.}} \ \ Before jointly learning the models, we pre-train the parser and the generator separately, using the labeled dataset, to enable the sampling of valid candidates with beam search when optimizing the lower bounds of dual information~(Eqs.~\ref{grad:1} and~\ref{grad:2}). The beam size is tuned from \{3,5\}. 
The parser and the generator are pre-trained until convergence. 
We also learn the language models for NL and MRs on the training sets beforehand, which are not updated during joint learning. Joint learning stops when the parser or the generator does not get improved for 5 continuous iterations. $\lambda$ is set to 0.1 for all the experiments. Additional descriptions about our setup are provided in Appendix~\ref{app:setup}. 

For the semi-supervised setup, since $\text{\scshape{Atis}}$ and $\text{\scshape{Django}}$ do not have additional unlabeled corpus and it is hard to obtain in-domain NL utterances and MRs, we create a new partial training set from the original training set via subsampling, and the rest is used as the unlabeled corpus. 
For $\text{\scshape{CoNaLa}}$, we subsample data from the full training set to construct the new training set and unlabeled set instead of sampling from the low-quality corpus which will much boost the data volume. 

\begin{table}[t]
\setlength{\abovecaptionskip}{0.2cm}
\centering
 \fontsize{8.5}{11.5}\selectfont
\begin{tabular}{lcccc}  
\toprule[.7pt]
 \multicolumn{5}{l}{\underline{$\text{\scshape{Semantic Parsing}}$ (in \textbf{Acc.})}}  \\ 
\textbf{Pro.}        & \scshape{Super}       & \scshape{DIM}      & \scshape{SemiDIM}     & \scshape{SelfTrain} \\
1/4         & 64.7         & 69.0       & \textbf{71.9}           & 66.3       \\
1/2         & 78.1         & 78.8       & \textbf{80.8}           & 79.2       \\
full        & 84.6         & \textbf{85.3}       & --             & --         \\ \hline 
\multicolumn{4}{l}{\underline{\textbf{\emph{Previous Supervised Methods}}}~(Pro. = full)} & \textbf{Acc.} \\
\multicolumn{4}{l}{$\text{\scshape{Seq2Tree}}$~\cite{dong-lapata:2016:P16-1}}     & 84.6       \\
\multicolumn{4}{l}{$\text{\scshape{ASN}}$~\cite{DBLP:conf/acl/RabinovichSK17}}        & 85.3       \\
\multicolumn{4}{l}{$\text{\scshape{ASN+SupATT}}$~\cite{DBLP:conf/acl/RabinovichSK17}} & 85.9      \\ 
\multicolumn{4}{l}{\textsc{Coarse2Fine}~\cite{DBLP:conf/acl/LapataD18}} & 87.7 \\
\toprule[.7pt] \toprule[.7pt]
\multicolumn{5}{l}{\underline{$\text{\scshape{NL Generation}}$ (in \textbf{BLEU})}} \\ 
\textbf{Pro.}  & \scshape{Super}  & \scshape{DIM}   & \scshape{SemiDIM}  & \scshape{BackBoost}  \\ 
1/4   & 36.9   & 37.7  & 39.1      & \textbf{40.9}        \\
1/2   & 39.1   & 40.7  & \textbf{40.9}      & 39.3        \\
full  & 39.3   & \textbf{40.6}  & --        & --           \\ \hline
\multicolumn{4}{l}{\underline{\textbf{\emph{Previous Supervised Methods}}}~(Pro. = full)} & \textbf{BLEU} \\ 
\multicolumn{4}{l}{$\text{\scshape{DeepCom}}$~\cite{DBLP:conf/iwpc/HuLXLJ18}} & 42.3 \\ \toprule[.7pt]
\end{tabular}
\caption{
Semantic parsing and NL generation results on $\text{\scshape{Atis}}$. {\bf Pro.}: proportion of the training samples used for training. Best result in each row is highlighted in bold. $|\text{full}| = $ 4,434.} 
\label{tab:atis}
\end{table}

\noindent{\textbf{Evaluation Metrics.}} Accuracy (\textit{Acc.}) is reported for parser evaluation based on exact match, and BLEU-4 is adopted for generator evaluation. For the code generation task in \textsc{CoNaLa}, we use BLEU-4 following the setup in \citet{DBLP:conf/msr/YinDCVN08}.

\noindent{\textbf{Baselines.}} We compare our methods of $\text{\scshape{DIM}}$ and $\text{\scshape{SemiDIM}}$ with the following baselines: 
\begin{compactitem}
	\item[\textbf{\textsc{Super}:}] Train the parser or generator separately without joint learning. The models for the parser and generator are the same as \textsc{DIM}. 
	\item[\textbf{\textsc{SelfTrain}}~\cite{lee2013pseudo}\textbf{:}] We use the pre-trained parser or generator to generate pseudo labels for the unlabeled \emph{sources}, then the constructed pseudo samples will be mixed with the labeled data to fine-tune the pre-trained parser or generator. 
	\item[\textbf{\textsc{BackBoost}}\textbf{:}] Adopted from the back translation method in \citet{DBLP:conf/acl/SennrichHB16}, which generates sources from unlabeled targets. The training process for \textsc{BackBoost} is the same as in \textsc{SelfTrain}. 
\end{compactitem} 
In addition to the above baselines, we also compare with popular supervised methods for each task, shown in the corresponding result tables. 

\begin{table}[t]
\setlength{\abovecaptionskip}{0.2cm}
\centering
 \fontsize{8.5}{11.5}\selectfont
\begin{tabular}{lcccc}  \toprule[.7pt]
 \multicolumn{5}{l}{\underline{$\text{\scshape{Code Generation}}$ (in \textbf{Acc.})}}    \\ 
\textbf{Pro.} & \scshape{Super} & \scshape{DIM} & \scshape{SemiDIM} & \scshape{BackBoost} \\ 
1/8  & 42.3  & 44.9 & \textbf{47.2}      & 47.0       \\
1/4  & 50.2  & 51.1 & \textbf{54.5}      & 51.7       \\
3/8  & 52.2  & 53.7 & 54.6      & \textbf{55.3}       \\
1/2  & 56.3  & 58.4 & \textbf{59.2}      & 58.9       \\ 
full & 65.1  & \textbf{66.6} &    --       & --          \\ \hline
\multicolumn{4}{l}{\underline{\textbf{\emph{Previous Supervised Methods}}}~(Pro. = full)} & \textbf{Acc.} \\
\multicolumn{4}{l}{LPN~\cite{DBLP:conf/acl/LingBGHKWS16}}   & 62.3       \\
\multicolumn{4}{l}{SNM~\cite{DBLP:conf/acl/YinN17}}   & 71.6      \\ 
\multicolumn{4}{l}{\textsc{Coarse2Fine}~\cite{DBLP:conf/acl/LapataD18}} & 74.1 \\
\toprule[.7pt] \toprule[.7pt]
\multicolumn{5}{l}{\underline{\textsc{Code Summarization} (in \textbf{BLEU})}} \\ 
\textbf{Pro.}   & \scshape{Super}   & \scshape{DIM}    & \scshape{SemiDIM}   & \scshape{SelfTrain}  \\ 
1/8    & 54.1    & 56.0   & \textbf{58.5}       & 54.4        \\
1/4    & 57.1    & 61.4   & \textbf{62.7}       & 58.0        \\
3/8    & 63.0    & 64.3   & \textbf{64.6}       & 63.0        \\
1/2    & 65.2    & 66.3   & \textbf{66.7}       & 65.4        \\
full   & 68.1    & \textbf{70.8}   & --         & --                \\ \hline
\multicolumn{4}{l}{\underline{\textbf{\emph{Previous Supervised Methods}}}~(Pro. = full)} & \textbf{BLEU} \\
\multicolumn{4}{l}{$\text{\scshape{DeepCom}}$~\cite{DBLP:conf/iwpc/HuLXLJ18}} & 65.9\\
\toprule[.7pt]
\end{tabular}
\caption{Code generation and code summarization results on $\text{\scshape{Django}}$. $|\text{full}| = $ 16,000.}
\label{tab:django}
\end{table}

\begin{table}
\setlength{\abovecaptionskip}{0.2cm}
\centering
 \fontsize{8.5}{11.5}\selectfont
\begin{tabular}{lcccc} \toprule[.7pt]
 \multicolumn{5}{l}{\underline{$\text{\scshape{Code Generation}}$ (in \textbf{BLEU})}}                                              \\ 
\textbf{Pro.}                 & \scshape{Super}          & \scshape{DIM}          & \scshape{SemiDIM}     & \scshape{BackBoost}     \\
1/2                  & 8.6            & \textbf{9.6}          & 9.5          & 9.0            \\
full                 & 11.1           & \textbf{12.4}         & --           & --             \\ 
\multicolumn{5}{l}{\underline{$\text{\scshape{Code Summarization}}$ (in \textbf{BLEU})}}                                        \\ 
\textbf{Pro.}                 & \scshape{Super}          & \scshape{DIM}          & \scshape{SemiDIM}     & \scshape{SelfTrain}     \\
1/2                  & 13.4           & 14.5         & \textbf{15.1}         & 12.7           \\
full                 & 22.5           & \textbf{24.8}         & --           & --             \\ \hline
\multicolumn{4}{l}{\underline{\emph{\textbf{Previous Supervised Methods}}}~(Pro. = full)} & \textbf{BLEU} \\ 
\multicolumn{4}{l}{\textsc{Code Gen.}: $\text{\scshape{NMT}}$~\cite{DBLP:conf/msr/YinDCVN08}} & 10.7 \\
\multicolumn{4}{l}{\textsc{Code Sum.}: $\text{\scshape{DeepCom}}$~\cite{DBLP:conf/iwpc/HuLXLJ18}}&  20.1 \\
\toprule[.7pt] \toprule[.7pt]
\multicolumn{5}{l}{\underline{After Pre-training (in \textbf{BLEU})}} \\ 
\multicolumn{1}{c}{} & \multicolumn{2}{c}{\underline{$\text{\scshape{Code Gen.}}$}} & \multicolumn{2}{c}{\underline{$\text{\scshape{Code Sum.}}$}} \\ 
\textbf{Pro.}                 & \scshape{Super}          & \scshape{DIM}          & \scshape{Super}        & \scshape{DIM}            \\
1/2                  & 10.3           & \textbf{10.6}         & \textbf{23.1}         & 23.0           \\
full                 & 11.1           & \textbf{12.4}         & 25.9         & \textbf{26.3}             \\ \hline
\multicolumn{4}{l}{\underline{\emph{\textbf{Previous Supervised Methods}}}~(Pro. = full)} & \textbf{BLEU} \\
\multicolumn{4}{l}{$\text{\scshape{Code Gen.}}$: $\text{\scshape{NMT}}$~\cite{DBLP:conf/msr/YinDCVN08}} & 10.9 \\
\multicolumn{4}{l}{$\text{\scshape{Code Sum.}}$: $\text{\scshape{DeepCom}}$~\cite{DBLP:conf/iwpc/HuLXLJ18}} & 26.5 \\
\toprule[.7pt]
\end{tabular}
\caption{Code generation and code summarization results on $\text{\scshape{CoNaLa}}$. For semi-supervised learning~(Pro. = 1/2), we sample 30K code snippets from the left data~(not used as training data) as unlabeled samples. $|\text{full}| = $ 90,000. 
}
\label{tab:conala}
\end{table}

\subsection{Results and Further Analysis}\label{exp:results}
\noindent{\textbf{Main Results with Full- and Semi-supervision.}}\ \ 
Results on the three datasets with supervised and semi-supervised setups are presented in Tables \ref{tab:atis}, \ref{tab:django}, and \ref{tab:conala}. 
For semi-supervised experiments on $\text{\scshape{Atis}}$, we use the NL part as extra unlabeled samples following~\citet{DBLP:conf/acl/NeubigZYH18}; for $\text{\scshape{Django}}$ and $\text{\scshape{CoNaLa}}$, unlabeled code snippets are utilized. 

We first note the consistent advantage of \textsc{DIM} over \textsc{Super} across all datasets and proportions of training samples for learning. This indicates that \textsc{DIM} is able to exploit the interaction between the dual tasks, and further improves the performance on both semantic parsing and NL generation.

For semi-supervised scenarios, \textsc{SemiDIM}, which employs unlabeled samples for learning, delivers stronger performance than \textsc{DIM}, which only uses labeled data. Moreover, \textsc{SemiDIM} outperforms both \textsc{SelfTrain} and \textsc{BackBoost}, the two semi-supervised learning methods. This is attributed to \textsc{SemiDIM}'s strategy of re-weighing pseudo samples based on the learning signals, which are indicative of their qualities, whereas \textsc{SelfTrain} and \textsc{BackBoost} treat all pseudo samples equally during learning. 
%
Additionally, we study the pre-training effect on \textsc{CoNaLa}. As can be seen in Table~\ref{tab:conala}, pre-training further improves the performance of \textsc{Super} and \textsc{DIM} on both code generation and summarization.

\begin{figure}[t]
\setlength{\abovecaptionskip}{0.2cm}
\begin{center}
\includegraphics[width=\columnwidth]{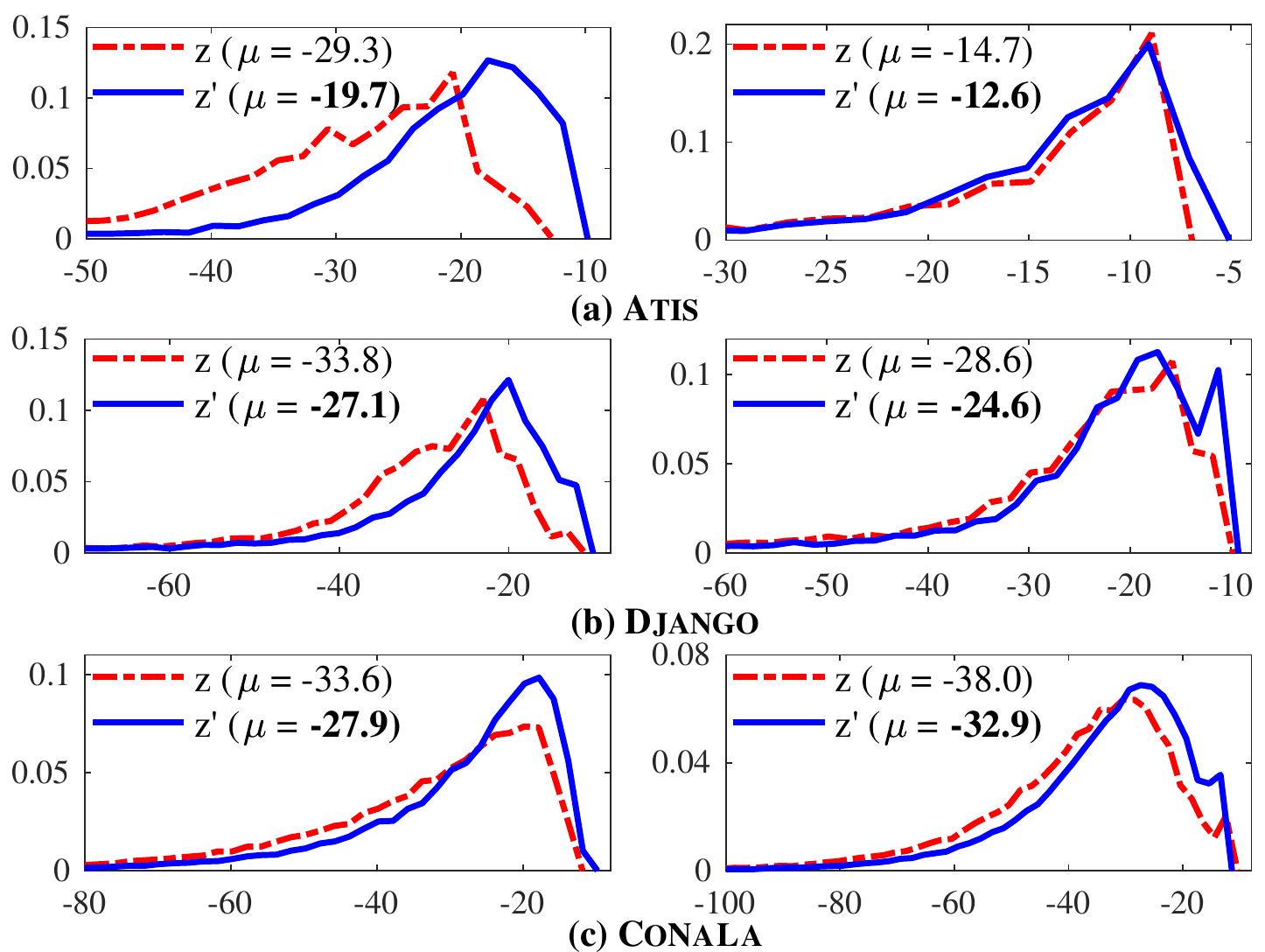}
\end{center}
   \caption{Lower bounds of the \emph{full} training set. $x$-axis: lower bound value; $y$-axis: frequency. The left column is for semantic parsing, and the right column for NL generation. $z$ is $\text{\scshape{Super}}$ method and $z'$ is $\text{\scshape{DIM}}$. $\mu$ is the average lower bound, with significantly better values boldfaced~($p < 0.01$).
   }
\label{fig:lower_bound}
\end{figure}

\begin{figure}[t]
\setlength{\abovecaptionskip}{0.2cm}
\begin{center}
\includegraphics[width=\columnwidth]{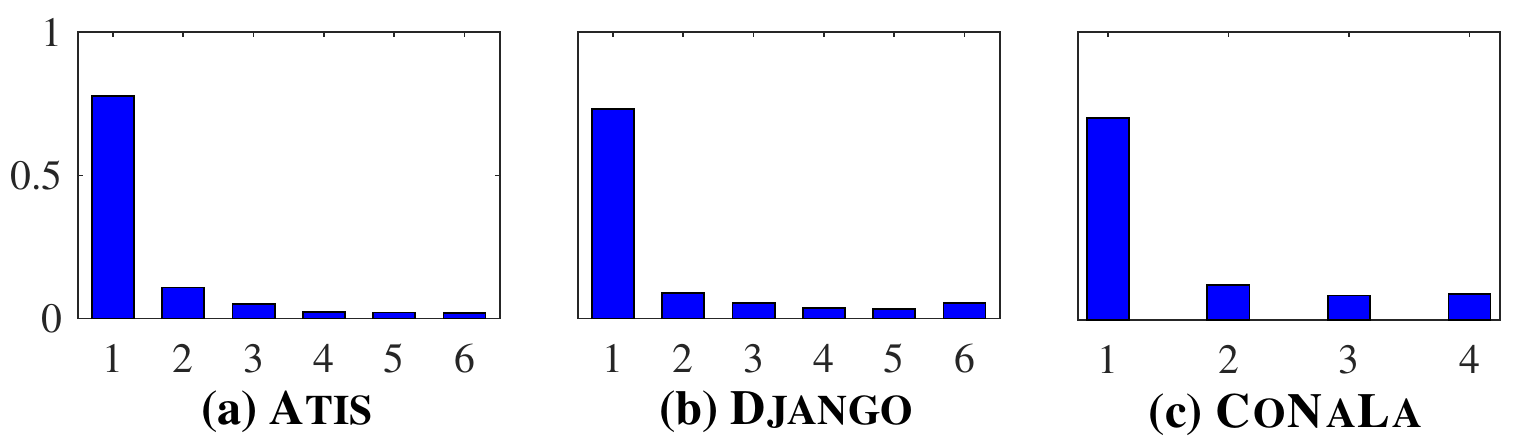}
\end{center}
   \caption{
   Distributions of the rank of learning signals over the gold-standard samples among the sampled set on unlabeled data using \textsc{SemiDIM}~(Pro. = 1/2).}
\label{fig:rank}
\end{figure}

\noindent{\textbf{Model Analysis.}}\ \ 
Here we study whether \textsc{DIM} helps enhance the lower bounds of the expected joint distributions of NL and MRs. Specifically, lower bounds are calculated as in Eqs.~\ref{dim:lower bound} and~\ref{eq:lower_bound_bw} on the full training set for models of $\text{\scshape{Super}}$ and $\text{\scshape{DIM}}$. As displayed in Fig.~\ref{fig:lower_bound}, $\text{\scshape{DIM}}$ better optimizes the lower bounds of both the parser and the generator, with significantly higher values of average lower bounds on the full data. These results further explains that when the lower bound of the primal model is improved, it produces learning signals of high quality for the dual model, leading to better performance on both tasks.

As conjectured above, $\text{\scshape{SemiDIM}}$ outperforms $\text{\scshape{SelfTrain}}$ in almost all setups because $\text{\scshape{SemiDIM}}$ re-weights the pseudo data with learning signals from the dual model. 
To demonstrate this, by giving the gold label for the unlabeled corpus, we rank the learning signal over the gold label among the sampled set using the semi-trained model, 
e.g. on \textsc{Atis}, given an NL $\mathbf{x}$ from the dataset used as the unlabeled corpus, we consider the position of the learning signal $l(\mathbf{x},\mathbf{y}^*;\phi)$ of gold-standard sample among all samples $\big \{l(\mathbf{x},\hat{\mathbf{y}}_i;\phi)|\hat{\mathbf{y}}_i \in S \big \}$. 
As seen in Fig.~\ref{fig:rank}, the gold candidates are almost always top-ranked, 
indicating that $\text{\scshape{SemiDIM}}$ is effective of separating pseudo samples of high and low-quality. 

\begin{table}[t]
\setlength{\abovecaptionskip}{0.2cm}
\centering
 \fontsize{8.5}{11.5}\selectfont
\begin{tabular}{p{2.3cm}ccc} \toprule[.5pt]
\scshape{Parser}    & \scshape{Super} & \lock~\scshape{Gen.}    & \scshape{DIM}  \\   \hline
\scshape{Atis}      & \graymid{84.6}  & \graylow{84.2}          & \grayhigh{85.3} \\
\scshape{Django}    & \graylow{65.1}  & \graymid{65.8}          & \grayhigh{66.6} \\
\scshape{CoNaLa}    & \graylow{11.1}  & \graymid{11.4}          & \grayhigh{12.4} \\ \toprule[.5pt] \toprule[.5pt]
\scshape{Generator} & \scshape{Super} & \lock~\scshape{Parser} & \scshape{DIM}  \\ \hline
\scshape{Atis}      & \graylow{39.3}  & \grayhigh{41.0}          & \graymid{40.6} \\
\scshape{Django}    & \graymid{68.1}  & \graylow{66.5}          & \grayhigh{70.8} \\
\scshape{CoNaLa}    & \graylow{22.5}  & \graymid{23.0}          & \grayhigh{24.8} \\ \toprule[.5pt]
\end{tabular}
\caption{Ablation study with full training set by freezing (\lock~) model parameters for generator or parser during learning. Darker indicates higher values.
}
\label{tab:ablation-study}
\end{table}

\noindent{\textbf{Ablation Study.}} \ \ 
We conduct ablation studies by training \textsc{DIM} with the parameters of parser or generator frozen. 
The results are presented in Table~\ref{tab:ablation-study}. As anticipated, for both of parsing and generation, when the dual model is frozen, the performance of the primal model degrades. 
This again demonstrates \textsc{DIM}'s effectiveness of jointly optimizing both tasks. Intuitively, jointly updating both the primal and dual models allows a better learned dual model to provide high-quality learning signals, leading to an improved lower bound for the primal. As a result, freezing parameters of the dual model has a negative impact on the learning signal quality, which affects primal model learning.


\begin{figure}[t]
\setlength{\abovecaptionskip}{0.2cm}
\begin{center}
\includegraphics[width=6.5cm]{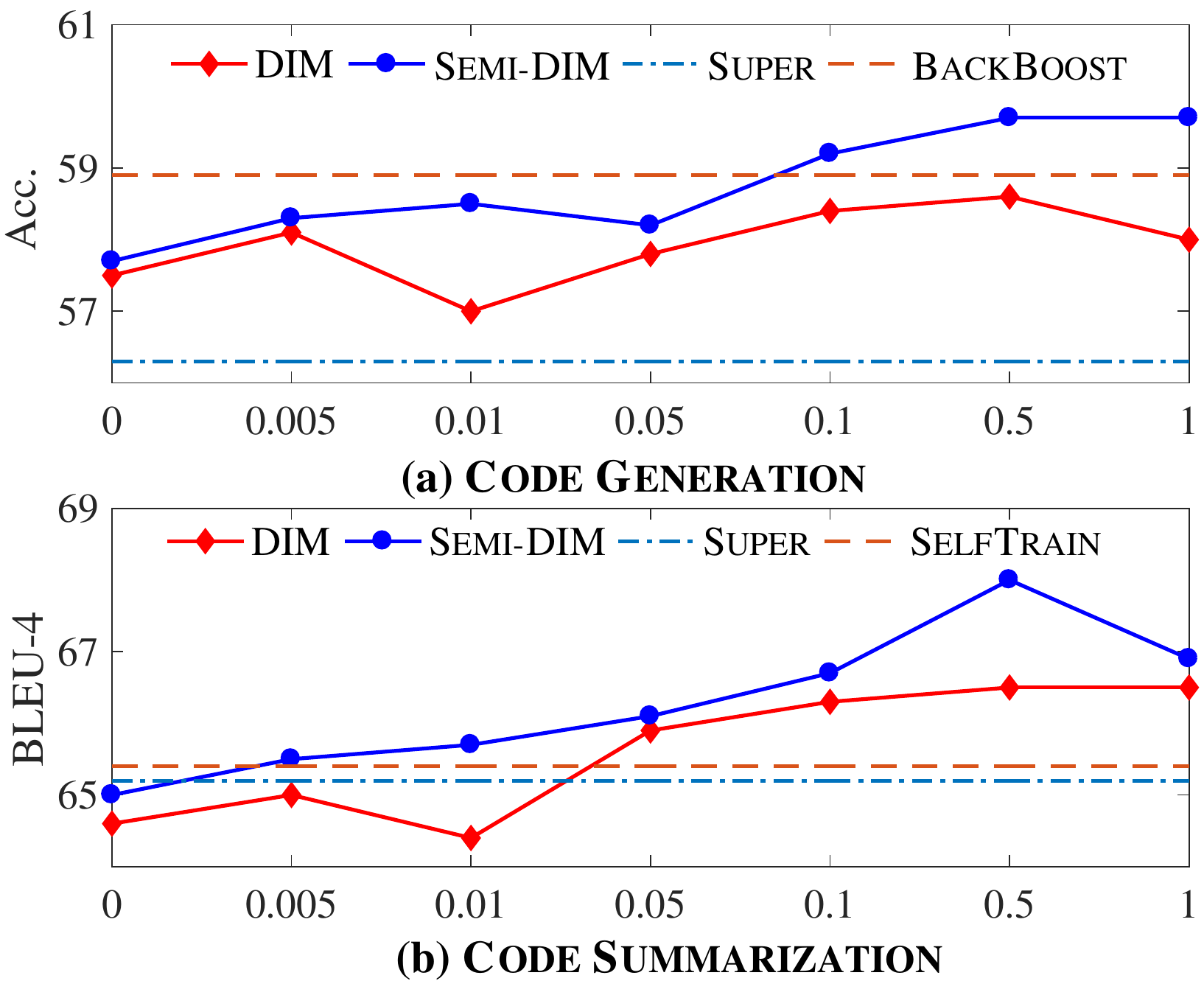}
\end{center}
   \caption{Model performance with different $\lambda$ values on $\text{\scshape{Django}}$~(Pro. = 1/2). }
\label{fig:lambda}
\end{figure}

\noindent{\textbf{Effect of $\lambda$.}} \ \ $\lambda$ controls the tradeoff between learning dual information and the unsupervised learning objective. 
Fig.~\ref{fig:lambda} shows that the optimal model performance can be obtained when $\lambda$ is within $0.1\sim1$. 
When $\lambda$ is set to $0$, the joint training only employs labeled samples, and its performance decreases significantly. 
A minor drop is observed at $\lambda=0.01$, which is considered to result from the variance of learning signals derived from the REINFORCE algorithm. 

\begin{figure}[t]
\setlength{\abovecaptionskip}{0.2cm}
\begin{center}
\includegraphics[width=7cm]{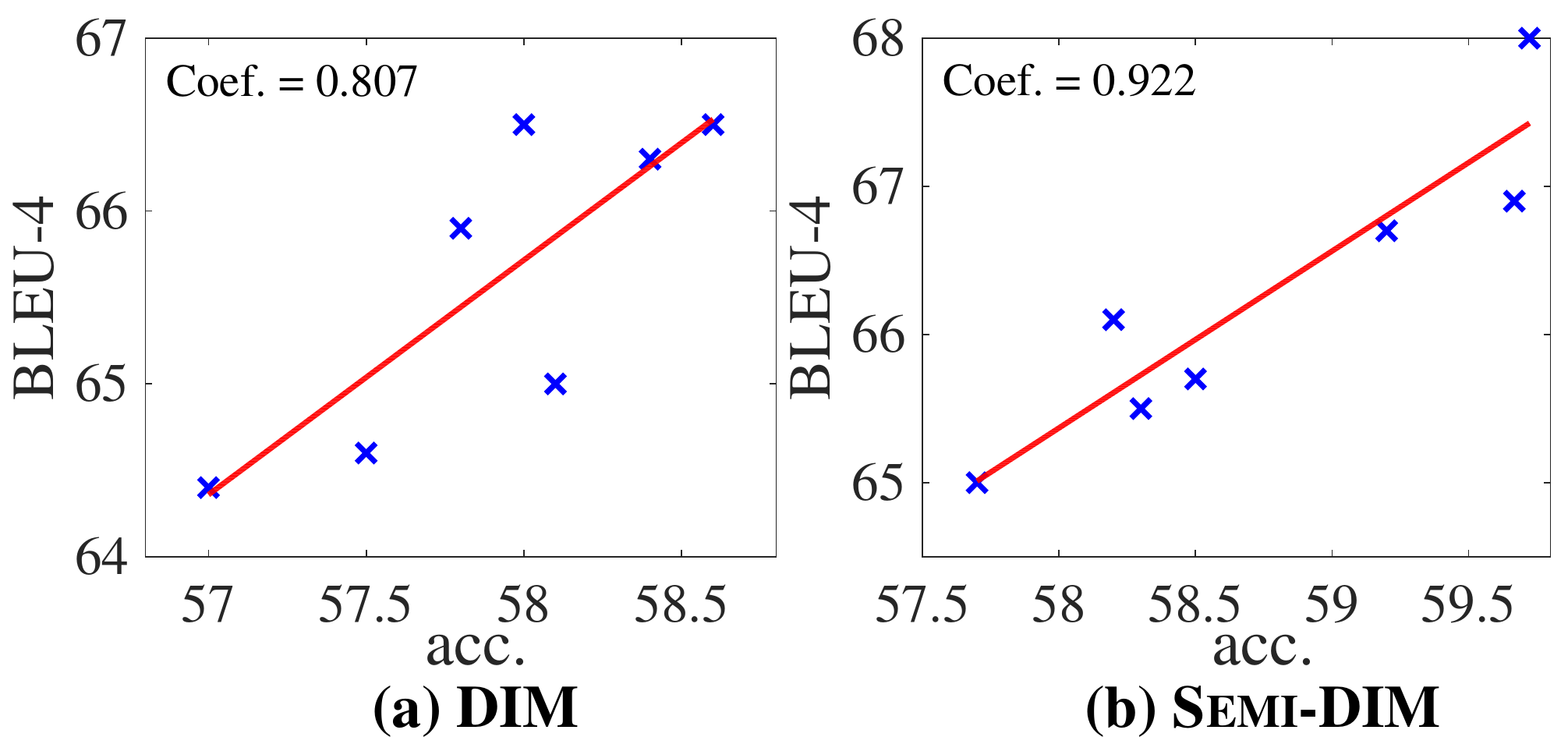}
\end{center}
   \caption{Performance correlation between parser and generator. $x$-axis is for parser and $y$-axis is for generator. Coef. indicates Pearson correlation coefficient. 
   }
\label{fig:coef}
\end{figure}

\noindent{\textbf{Correlation between Parser and Generator.}} \ \ 
We further study the performance correlation between the coupled parser and generator. 
Using the model outputs shown in Fig.~\ref{fig:lambda}, we run linear regressions of generator performance on parser performance, and a high correlation is observed between them (Fig.~\ref{fig:coef}). 


\section{Related Work}
\noindent{\textbf{Semantic Parsing and NL Generation.}} Neural sequence-to-sequence models have achieved promising results on semantic parsing~\cite{dong-lapata:2016:P16-1,DBLP:conf/acl/JiaL16,DBLP:conf/acl/LingBGHKWS16,DBLP:conf/acl/LapataD18} and natural language generation~\cite{DBLP:conf/acl/IyerKCZ16,DBLP:conf/acl/KonstasIYCZ17,DBLP:conf/iwpc/HuLXLJ18}. To better model structured MRs, tree structures and more complicated graphs are explored for both parsing and generation~\cite{dong-lapata:2016:P16-1,DBLP:conf/acl/RabinovichSK17,DBLP:conf/acl/YinN17,DBLP:conf/acl/GildeaWZS18,DBLP:conf/acl/ChengRSL17,DBLP:journals/corr/abs-1808-01400}. 
Semi-supervised learning has been widely studied for semantic parsing~\cite{DBLP:conf/acl/NeubigZYH18,DBLP:conf/emnlp/KociskyMGDLBH16,DBLP:conf/acl/JiaL16}. 
Similar to our work, \citet{DBLP:conf/kbse/ChenZ18} and \citet{DBLP:conf/icml/AllamanisTGW15} study code retrieval and code summarization jointly to enhance both tasks. Here, we focus on the more challenging task of code generation instead of retrieval, and we also aim for general-purpose MRs. 

\noindent{\textbf{Joint Learning in NLP.}} \ \ There has been growing interests in leveraging related NLP problems to enhance primal tasks~\cite{DBLP:journals/jmlr/CollobertWBKKK11,DBLP:conf/acl/PengTS17,DBLP:conf/aaai/LiuLZS16}, e.g. sequence tagging~\cite{DBLP:journals/jmlr/CollobertWBKKK11}, dependency parsing~\cite{DBLP:conf/acl/PengTS17}, discourse analysis~\cite{DBLP:conf/aaai/LiuLZS16}. 
Among those, multi-task learning~(MTL)~\cite{DBLP:journals/jmlr/AndoZ05} is a common method for joint learning, especially for neural networks where parameter sharing is utilized for representation learning. 
We 
follow the recent work on dual learning~\cite{DBLP:conf/icml/XiaQCBYL17} to train dual tasks, where interactions can be employed to enhance both models. 
Dual learning has been successfully applied in NLP and computer vision problems, such as neural machine translation~\cite{DBLP:conf/nips/HeXQWYLM16}, question generation and answering~\cite{DBLP:journals/corr/TangDQZ17}, image-to-image translation~\cite{DBLP:conf/iccv/YiZTG17,DBLP:conf/iccv/ZhuPIE17}. Different from~\citet{DBLP:conf/icml/XiaQCBYL17} which minimizes the divergence between the two expected joint distributions, we aim to learn the expected distributions in a way similar to distribution matching~\cite{DBLP:conf/nips/GanCWPZLLC17}. Furthermore, our method can be extended to semi-supervised scenario, prior to \citet{DBLP:conf/icml/XiaQCBYL17}'s work which can only be applied in supervised setup. Following \citet{DBLP:conf/nips/ZhangGGGLBD18}, we deduce the variational lower bounds of expected distributions via information maximization~\cite{DBLP:conf/nips/BarberA03}. DIM aims to optimize the dual information instead of the two mutual information studied in \citet{DBLP:conf/nips/ZhangGGGLBD18}.

\section{Conclusion}
In this work, we propose to jointly train the semantic parser and NL generator by exploiting the structural connections between them. We introduce the method of \textsc{DIM} to exploit the duality, and provide a principled way to optimize the dual information. We further extend supervised \textsc{DIM} to semi-supervised scenario~(\textsc{SemiDIM}). Extensive experiments demonstrate the effectiveness of our proposed methods. 

To overcome the issue of poor labeled corpus for semantic parsing, some automatically mined datasets have been proposed, e.g. \textsc{CoNaLa}~\cite{DBLP:conf/msr/YinDCVN08} and \textsc{StaQC}~\cite{DBLP:conf/www/YaoWCS18}. However, these datasets are noisy and it is hard to train robust models out of them. In the future, we will further apply DIM to learn semantic parser and NL generator from the noisy datasets.

\section*{Acknowledgments}
The work described in this paper is supported by Research Grants  Council  of  Hong  Kong  (PolyU  152036/17E) and National Natural Science Foundation of China (61672445). Lu Wang is supported by National Science Foundation through Grants IIS-1566382 and IIS-1813341. This work was done when Hai was a research assistant in PolyU from Oct. 2018 to March 2019.  

\bibliography{DIM}
\bibliographystyle{acl_natbib}

\newpage
\clearpage
\appendix

\section{Model Details for the Parser and Generator}\label{app:framework}
The parser and generator have the same seq2seq framework. We take the parser for example. Given the NL utterance $\mathbf{x}$ and the linearized MR $\mathbf{y}$, we use bi-LSTM to encode $\mathbf{x}$ into context vectors, and then a LSTM decoder generates $\mathbf{y}$ from the context vectors. The parser $p_\theta(\mathbf{y}|\mathbf{x})$ is formulated as following: 
\begingroup
\setlength\abovedisplayskip{4pt}\setlength\belowdisplayskip{4pt}
\begin{equation}
p_\theta(\mathbf{y}|\mathbf{x}) = \prod_{t=1}^{|\mathbf{y}|} p_\theta(y_t|\mathbf{y}_{<t}, \mathbf{x})
\label{overall_eq}
\end{equation}
\endgroup
where $\mathbf{y}_{<t} = y_1 \cdots y_{t-1}$. 

The hidden state vector at time $t$ from the encoder is the concatenation of forward hidden vector $\overrightarrow{\mathbf{h}}_t$ and backward one $\overleftarrow{\mathbf{h}}_t$, denoted as $\mathbf{h}_t = [\overrightarrow{\mathbf{h}}_t, \overleftarrow{\mathbf{h}}_t]$. With the LSTM unit $f_{\text{LSTM}_{\text{e}}}$ from the encoder, we have $\overrightarrow{\mathbf{h}}_t = f_{\text{LSTM}_{\text{e}}}(x_t, \overrightarrow{\mathbf{h}}_{t-1})$ and $\overleftarrow{\mathbf{h}}_t = f_{\text{LSTM}_{\text{e}}}(x_t, \overleftarrow{\mathbf{h}}_{t+1})$. 
 
From the decoder side, using the decoder LSTM unit $f_{\text{LSTM}_{\text{d}}}$,  we have the hidden state vector at time $t$ as $\mathbf{s}_t = f_{\text{LSTM}_{\text{d}}}(y_{t-1}, \mathbf{s}_{t-1})$. 
Global attention mechanism~\cite{DBLP:conf/emnlp/LuongPM15} is applied to obtain the context vector $\mathbf{c}_t$ at time $t$:
\begingroup
\setlength\abovedisplayskip{4pt}\setlength\belowdisplayskip{4pt}
\begin{equation}
\nonumber \mathbf{c}_t =  \sum_{i=1}^{|\mathbf{x}|}\alpha_{t,i}\mathbf{h}_i
\end{equation}
\endgroup
where $\alpha_{t,i}$ is the attention weight and is specified as:
\begingroup
\setlength\abovedisplayskip{4pt}\setlength\belowdisplayskip{4pt}
\begin{equation}
    \alpha_{t,i} =  \frac{ \exp(\mathbf{W}_{\text{att}}[\mathbf{s}_t;\mathbf{h}_i])}{\sum_{k=1}^{|x|}\exp(\mathbf{W}_{\text{att}}[\mathbf{s}_t \mathbf{h}_k])}
\end{equation}
\endgroup
where $\mathbf{W}_{\text{att}}$ is the learnable parameters. 

At time $t$, with hidden state $\mathbf{s}_t$ in the decoder and context vector $\mathbf{c}_t$ from the encoder, we have the prediction probability for $y_t$:
\begingroup
\setlength\abovedisplayskip{4pt}\setlength\belowdisplayskip{4pt}
\begin{align}
\nonumber p_{\text{vocab}}(y_t|\mathbf{y}_{<t}, \mathbf{x}) = \mathbf{f}_{\text{softmax}}(\mathbf{W}_{\text{d}_\text{1}} \cdot 
\tanh (\mathbf{W}_{\text{d}_\text{2}}[\mathbf{s}_t;\mathbf{c}_t]) )
\end{align}
\endgroup
where $\mathbf{W}_{\text{d}_\text{1}}$ and $\mathbf{W}_{\text{d}_\text{2}}$ are learnable parameters. 

We further apply the pointer-network~\cite{DBLP:conf/nips/VinyalsFJ15} to copy tokens from the input to alleviate the out-of-vocabulary~(OOV) issue. We adopt the calculation flows for copying mechanism from \citet{DBLP:conf/acl/NeubigZYH18}, readers can refer to that paper for further details. 

\section{Marginal Distributions}\label{app:marginal}
To estimate the marginal distributions $p(\mathbf{x})$ and $q(\mathbf{y})$, we learn the LSTM language models over the NL utterances and MRs. MRs are linearized. Suppose given the NL $\mathbf{x} = \{x_i\}_{i=1}^{|\mathbf{x}|}$, the learning objective is:
\begingroup
\setlength\abovedisplayskip{4pt}\setlength\belowdisplayskip{4pt}
\begin{equation}
    p(\mathbf{x}) = \prod_{i=1}^{|\mathbf{x}|}p(x_i|x_{<i})
\end{equation}
\endgroup
where $x_{<i} = x_1 \cdots x_{i-1}$. At time $t$, we have the following probability to predict $x_t$:
\begingroup
\setlength\abovedisplayskip{4pt}\setlength\belowdisplayskip{4pt}
\begin{equation}
    p(x_t | x_{<t}) = f_{\text{softmax}}(\mathbf{W}\cdot\mathbf{h}_t+\mathbf{b})
\end{equation}
\endgroup
Here, $\mathbf{h}_t$ is estimated using the LSTM network:
\begingroup
\setlength\abovedisplayskip{4pt}\setlength\belowdisplayskip{4pt}
\begin{equation}
    \mathbf{h}_t = f_{\text{LSTM}}(x_t, \mathbf{h}_{t-1})
\end{equation}
\endgroup

The above marginal distribution estimation for NLs is also applied to linearized MRs. 

\section{Experimental Setups}\label{app:setup}
\subsection{Marginal Distribution}
We pre-train the language models on the full training set before joint learning and the language mdoels will be fixed in the following experiments. The embedding size is selected from $\{128, 256\}$ and the hidden size is tuned from $\{256, 512\}$, which are both evaluated on the validation set. We use SGD to update the models. Early stopping is applied and the training will be stopped if the ppl value does not decrease for continuous $5$ times. 

\subsection{Model Configuration}
To conduct the joint learning using \textsc{DIM} and \textsc{SemiDIM}, we have to firstly train the parser and generator separately referred as the method of \textsc{Super}. 

To pre-train the parser and generator, we tune the embedding size from $\{125, 150, 256\}$ and hidden size from $\{256, 300, 512\}$. The batch size is selected from $\{10, 16\}$ varying over the datasets. Early stopping is applied and the patience time is set to $5$. Initial learning rate is $0.001$. Adam is adopted to optimize the models. The parser and generator will be trained until convergence. 

After the pre-training, we conduct joint learning based on the pre-trained parser and generator. The learning rate will be slowed down to $0.00025$. The beam size for sampling is tuned from $\{3,5\}$.

\end{document}